\pdfoutput=1

\documentclass[11pt]{article}

\usepackage{ACL2023}
\usepackage{graphicx}
\usepackage{amsmath}
\usepackage{amsfonts}
\usepackage{xurl} 
\usepackage{times}
\usepackage{latexsym}
\usepackage{booktabs} 
\usepackage{caption}  

\usepackage{listings}

\usepackage[T1]{fontenc}

\usepackage[utf8]{inputenc}

\usepackage{inconsolata}
\usepackage{pifont}     
\newcommand{\cmark}{\textcolor{green!60!black}{\ding{51}}}
\newcommand{\xmark}{\textcolor{red}{\ding{53}}}

\usepackage{mathtools}
\usepackage{pgfplots}
\usepackage{pgfplotstable}
\usepackage{adjustbox}
\pgfplotsset{compat=1.18}
\usepackage{pgfplots}
\usepackage{pgfplotstable}
\pgfplotsset{compat=1.18}
\usepackage{sansmath}
\usepackage{pgf-pie}  
\usepackage{tikz}
\usepackage{tcolorbox}
\usepackage{xcolor}
\definecolor{blue2}{RGB}{112,173,220}
\definecolor{blue3}{RGB}{91,206,250}
\definecolor{pink2}{RGB}{245,169,184}
\definecolor{green2}{RGB}{180,210,124}
\usepackage[activate=false,kerning]{microtype}
\SetExtraKerning[unit=character]{encoding=*}{\textemdash={100,100}}

\usepackage{colortbl}

\definecolor{highlightgreen}{HTML}{E5F5E0}
\title{\textsc{PhysicsEval}: Inference-Time Techniques to Improve the Reasoning Proficiency of Large Language Models on Physics Problems}

\author{\bf Oshayer Siddique$^*$, 
{\bf J. M Areeb Uzair Alam$^*$,}
{\bf Md Jobayer Rahman Rafy$^*$,}\\
{\bf Syed Rifat Raiyan$^{*\text{\textdagger}}$,} 
{\bf Hasan Mahmud,}
{\bf Md Kamrul Hasan}\\
Systems and Software Lab (SSL)\\Department of Computer Science and Engineering\\
Islamic University of Technology, Dhaka, Bangladesh\\
\texttt{\small\{oshayer, areebuzair, jobayerrahman, rifatraiyan, hasan, hasank\}@iut-dhaka.edu}\\
}
\newcommand{\physicseval}{\textsc{PhysicsEval}}
\newcommand{\physicsevals}{\textsc{PhysicsEval} }
\begin{document}
\maketitle
\def\thefootnote{*}\footnotetext{Equal contribution}
\def\thefootnote{\textdagger}\footnotetext{Corresponding author}
\def\thefootnote{\arabic{footnote}}
\begin{abstract}
The discipline of physics stands as a cornerstone of human intellect, driving the evolution of technology and deepening our understanding of the fundamental principles of the cosmos. Contemporary literature includes some works centered on the task of solving physics problems---a crucial domain of natural language reasoning. In this paper, we evaluate the performance of frontier LLMs in solving physics problems, both mathematical and descriptive. We also employ a plethora of inference-time techniques and agentic frameworks to improve the performance of the models. This includes the verification of proposed solutions in a cumulative fashion by other, smaller LLM agents, and we perform a comparative analysis of the performance that the techniques entail. 
There are significant improvements when the multi-agent framework is applied to problems that the models initially perform poorly on. Furthermore, we introduce a new evaluation benchmark for physics problems, \physicseval, consisting of 19,609 problems sourced from various physics textbooks and their corresponding correct solutions scraped from physics forums and educational websites. Our code and data are publicly available at \url{https://github.com/areebuzair/PhysicsEval}.
\end{abstract}

\section{Introduction}
In the preliminary stages of development, LLMs predominantly exhibited a spuriously correlated reliance on rapid, intuitive processing. As per the dual-process theory by \citet{kahneman2011thinking}---that posits a fundamental dichotomy between fast, intuitive (System 1) and slower, deliberate (System 2) modes of thought---the inherent System 1 bias of nascent LLMs presented a fundamental weakness in their ability to engage in complex, multi-step reasoning. With the advent of prompting techniques such as Chain-of-Thought (CoT) and Tree-of-Thought (ToT), LLMs have bridged this gap and made significant improvements in logical reasoning tasks \cite{wei2023chainofthoughtpromptingelicitsreasoning,yao2023tree}. 
\begin{figure}[t]
\centering
\begin{tcolorbox}[boxsep=1pt,left=5pt,right=5pt,colback=blue2!5!white, colframe=blue3, title=Problem:]
\footnotesize
At what rate does the Sun emit photons? For simplicity, assume that the Sun’s entire emission at the rate of $3.9\times 10^{26}\operatorname{W}$ is at the single wavelength of $550 \operatorname{nm}$.
\end{tcolorbox}
\vspace{-2mm}
\begin{tcolorbox}[boxsep=1pt,left=5pt,right=5pt,colback=green2!5!white, colframe=green2, title=Solution:]
\small
    \textbf{1. Describe the expression of photon energy.}\\
The energy \( E \) of a photon of wavelength \( \lambda \) is given by,
\vskip -1.4\abovedisplayskip
\begin{align*}
E = \frac{h c}{\lambda}
\end{align*}
\vskip -0.8\abovedisplayskip

Here, \( h \) is Planck’s constant, and \( c \) is the speed of light.
    
    \vspace{2mm}
    \textbf{2. Determine the rate of emission of the photon.}\\
Assume that the photons are emitted by a rate \( R \) from the sodium lamp. Then, the power \( P \) of the sodium lamp is equal to the product of rate \( R \) and the energy of each photon \( E \).
\vskip -1.7\abovedisplayskip
\begin{align}
\label{eq:example}
P = R E &\Rightarrow
P = R \frac{h c}{\lambda}
\Rightarrow R = \frac{P \lambda}{h c}
\end{align}
\vskip -0.8\abovedisplayskip

Substitute the below values in Equation \ref{eq:example}.\\
\vskip -0.6\abovedisplayskip
\( P = 3.9 \times 10^{26} \, \operatorname{W} \)\\
\( \lambda = 550 \, \operatorname{nm} \)\\
\( h = 6.626 \times 10^{-34} \, \operatorname{Js} \)\\
\( c = 3 \times 10^8 \, \operatorname{m/s} \)
\vspace{1mm}

Therefore, the rate of emitted photons from the Sun is:
\vskip -1.4\abovedisplayskip
\begin{align*}
R &= \frac{(3.9 \times 10^{26} \, \operatorname{W}) \cdot (550 \times 10^{-9} \, \operatorname{m})}{(6.626 \times 10^{-34} \, \operatorname{Js}) \cdot (3.00 \times 10^{8} \, \operatorname{m/s})}\\
&= \boxed{1.08 \times 10^{45} \, \operatorname{photons/s}}
\end{align*}
\end{tcolorbox}
\caption{Example of an astrophysics problem from the \physicsevals benchmark.}
\end{figure}
Despite substantial progress in mathematical problem solving, large language models continue to face significant challenges in accurately reasoning through physics problems. Google's generative AI chatbot Bard, based on LaMDA \cite{thoppilan2022lamda} and PaLM \cite{chowdhery2023palm} models, ranked in the bottom $2\%$ of test takers on the Physics GRE \cite{gupta2023testingllmperformancephysics}.

Most research works in recent literature pertaining to improved performance of LLMs in physics include specialized training techniques, such as the method proposed by \citet{anand2024enhancingllmsphysicsproblemsolving}. 
These approaches primarily falter because the LLMs often commit mishaps during their reasoning process, including calculation mistakes, misinterpretation of physical scenarios, and dimensional inconsistencies \citep{ding2023usinglargelanguagemodel}. However, upon adoption of proper prompting heuristics, these mistakes are susceptible to being detected during inference time by LLMs \cite{pang-etal-2025-physics}. This necessitates either a self-correcting framework or an agentic framework with checks and balances that can detect and reconcile the corrigible aspects of a solver model's response. To this end, we utilize a self-refinement technique, where the solver LLM checks its own answer once before generating the final response. In the same vein, we build upon established agentic paradigms by systematically adapting and evaluating these frameworks for the specific, complex domain of physics. We also explore a novel orchestration with a separate, smaller group of LLM agent(s), to determine their efficacy in identifying and correcting domain-specific errors.
It aims to reduce computational overhead while providing an unbiased assessment, addressing the tendency of LLMs to accept their own outputs as correct unless explicitly fine-tuned otherwise \citep{kadavath2022languagemodelsmostlyknow}. It also enables the use of commercial LLMs for problem-solving while delegating verification to open-source models, thereby minimizing financial overhead via API usage.

In order to facilitate rigorous evaluation of physics reasoning in LLMs, we introduce \physicseval, an evaluation benchmark comprising 19,609 physics problems and their elaborated solutions. This dataset is curated by scraping problems and initial solutions from online educational forums, followed by a polishing phase via Google's Gemini 2.5 Pro \cite{google2025gemini2_5pro} to generate detailed, step-by-step explanations. From this collection, we randomly sample $10\%$ ($\approx$ 1,962 problems) to assess the performance of frontier LLMs across four inference-time techniques: baseline inference, self-correction, single-verifier cumulative reasoning, and a multi-agent review framework. The remaining problems are reserved for future model training or fine-tuning efforts.
\section{Related Work}
The natural language reasoning domain is now confronted with the non-trivial problem, which is that the established mathematical benchmarks are proving to be insufficient. As foundation models like OpenAI's o1 \cite{jaech2024openai} begin to master these evaluations---a challenge anticipated by \citet{phan2025humanity} who proposed Humanity’s Last Exam---it becomes important to distinguish between mere pattern matching and genuine scientific reasoning. Consequently, the field must evolve toward better evaluation benchmarks, not simply to measure capability, but to compel these models to grapple with the kind of difficult problems that actually constitute meaningful progress. The evaluation of LLMs in mathematical reasoning has been characterized by a progressive escalation in benchmark difficulty, driven by advancing model capabilities. Foundational datasets like MAWPS \cite{koncel2016mawps}, SVAMP \cite{patel2021nlp}, ParaMAWPS \cite{raiyan-etal-2023-math}, GSM8K \cite{cobbe2021gsm8k}, and MATH \cite{hendrycks2measuring}, which cover up to undergraduate-level content, have been largely surmounted by contemporary models. This performance plateau has spurred the creation of a new tier of challenges. More recent benchmarks, including HARDMath \cite{fan24hardmath}, OlympiadBench \cite{he2024olympiadbench}, PutnamBench \cite{tsoukalas2024putnambench}, and FrontierMath \cite{glazer2024frontiermath}, represent this next frontier, introducing complex university and olympiad-level problems to continue probing the limits of the most advanced systems.
\begin{table*}[t]
    \centering
    \resizebox{0.89\textwidth}{!}{%
    \begin{tabular}{l c c cc ccc}
        \toprule
        \textbf{Benchmark} & \textbf{Size} & \textbf{Knowledge} & \multicolumn{2}{c}{\textbf{Question}} & \multicolumn{3}{c}{\textbf{Solution}} \\
        \cmidrule(lr){4-5} \cmidrule(lr){6-8}
        & & & \textbf{Type} & \textbf{Avg. Tokens} & \textbf{Step-by-step} & \textbf{Avg. Tokens} & \textbf{Avg. Steps} \\
        \midrule
        JEEBench       & 123   & CEE     & OE, MC & 169.7 & \xmark      & $-$      & $-$   \\
        MMLU-Pro       & 1,299  & COL     & MC    & 52.1  & \cmark      & $-$      & $-$   \\
        GPQA           & 227   & PH.D.   & OE    & 111.4 & \xmark & 197.2  & 3.6 \\
        SciEval        & 1,657  & $-$       & OE, MC & 154.5 & \xmark      & $-$      & $-$   \\
        SciBench       & 295   & COL     & OE    & 80.5  & \xmark & 315.9  & 2.8 \\
        MMMU           & 443   & COL     & OE, MC & 53.8  & \xmark      & $-$      & $-$   \\
        ScienceQA      & 617   & K1-K12  & MC    & 13.3  & \xmark & 63.0   & 2.4 \\
        OlympiadBench  & 2,334  & COMP    & OE    & 222.0 & \xmark & 199.8  & 3.7 \\
        EMMA           & 156   & $-$       & MC    & 109.5 & \xmark      & $-$      & $-$   \\
        PhysReason           & 1,200   & CEE+COMP       & OE    & 226.3 & \cmark      & 441.3      & 8.1   \\
        UGPhysics           & 11,040   & COL       & OE, MC    & 82.4 & \cmark      & 318.5      & $-$   \\
        \midrule
        \physicseval           & 19,609   & CEE+COL+COMP       & OE    & 98.8 & \cmark      & 3830.8      & 3.9   \\
        \bottomrule
    \end{tabular}
    } 
    \caption{Comparison of \physicsevals with various other physics reasoning benchmarks. For the \textbf{"Knowledge"} column: COMP refers to Competition level, COL to College level, CEE to College Entrance Examination, K1-K12 to elementary and high school levels, and PH.D. to Doctor of Philosophy. For the \textbf{"Question Type"} column: OE denotes open-ended questions, while MC stands for multiple-choice questions.}
    \label{tab:dataset-comparison}
\end{table*}

Likewise, it is necessary to move beyond \textit{ad hoc} mathematical problem-solving to evaluate reasoning grounded in the physical world. The domain of physics requires models to integrate an understanding of fundamental laws and real-world principles with the capacity for complex, multi-step deductions. Consequently, there is a clear imperative for specialized benchmarks tailored to the unique complexities of advanced physics reasoning. The evolution of physics-reasoning benchmarks for LLMs has progressed systematically from rudimentary problem collections to sophisticated assessment frameworks that evaluate multidimensional aspects of physics comprehension. Initial efforts, exemplified by PhyQA \cite{ding2023using} and UGPhysics \cite{xu2025ugphysics}, established foundational benchmark suites comprising thousands of structured introductory problems purported to assess basic conceptual understanding. Subsequently, more advanced evaluation instruments such as PhysBench \cite{qiu2025phybench} and PhysReason \cite{zhang2025physreason} introduced complex problem formulations requiring extended reasoning chains and multi-step analytical processes. Contemporary benchmark development has culminated in research-oriented assessment suites such as TP-Bench \cite{chung2025theoretical} and CURIE \cite{cui2025curie}, multi-modal frameworks like MMPhyQA \cite{anand2024mmphyqa} that incorporate visual reasoning components, and specialized domain-specific instruments including FEABench \cite{mudur2025feabench}.

With the intent of eliciting sound reasoning in LLMs for problem-solving, researchers have proposed various methods that try to emulate the thought processes of humans. Cumulative Reasoning (CR) by \citet{zhang2023cumulative}, in an iterative fashion involving proposer and verifier LLMs, decomposes problems into smaller and more manageable subproblems and utilizes premises from previous iterations to enhance LLM reasoning. Other works explore self-improvement through verification feedback for better LLM capabilities across diverse domains, including reasoning \cite{shinn2023reflexion,hong-etal-2024-closer} and security \cite{li2024deepinception,pmlr-v235-cao24d}. Specifically for physics reasoning, Physics Reasoner \cite{pang-etal-2025-physics} adopts knowledge-augmentation to facilitate germane formula retrieval, in order to have a properly guided reasoning stage. They also find that the incorporation of checklists acts as a good buttress for LLMs' self-improvement.

\section{\physicsevals Benchmark}
To test our inference-time techniques, we curate a new dataset of physics problems, namely \physicseval. We compile problems and solutions from various textbooks, spanning from high school to university levels. The list of physics books from which we source the problems is provided in the Appendix \ref{app:books}. We then use the Gemini 2.5 Pro model by Google\footnote{\url{https://blog.google/technology/google-deepmind/gemini-model-thinking-updates-march-2025/}} to elaborate the solutions into logical steps and sub-steps. The core content of both the problems and solutions in \physicsevals originates entirely from human-generated sources. Problems are scraped from established physics textbooks, which are listed in Appendix \ref{app:books}. The corresponding solutions are scraped from reputable online physics forums and educational websites where experts and educators provide answers. These scraped texts often contain grammatical errors, typos, inconsistent formatting, and poorly rendered equations. Gemini 2.5 Pro is employed solely in a \textit{"polishing"} capacity to clean and standardize this content. Its role is strictly limited to correcting grammar, fixing typos, properly formatting \LaTeX\;equations, and organizing the existing human reasoning into a clear, step-by-step structure. This process does not involve adding, altering, or augmenting the substantive scientific content. The term "elaborate" alludes to this syntactic reformatting and structuring process---which makes the human-written solutions more readable and consistent---not to generating \textit{de novo} explanations. This ensures the ground truth remains faithful to human expertise and avoids introducing model-specific biases. The categories of the problems and their key topics are also extracted. Consequently, we are left with a dataset of size 19,609. We partition this dataset into a train set and a test set, maintaining a 90:10 split.

Table \ref{tab:dataset-comparison} depicts a statistical comparison of \physicsevals with the physics reasoning benchmarks in the existing literature. For the sake of brevity, a complete description of \physicseval, including sources, format, examples, comparisons with other datasets, statistics, etc., can be found in Appendix \ref{app:dataset}.

\section{Methodology}
\subsection{Multi-Agent Review Framework}
\begin{figure*}[h]
    \centering
    \includegraphics[width=0.84\textwidth]{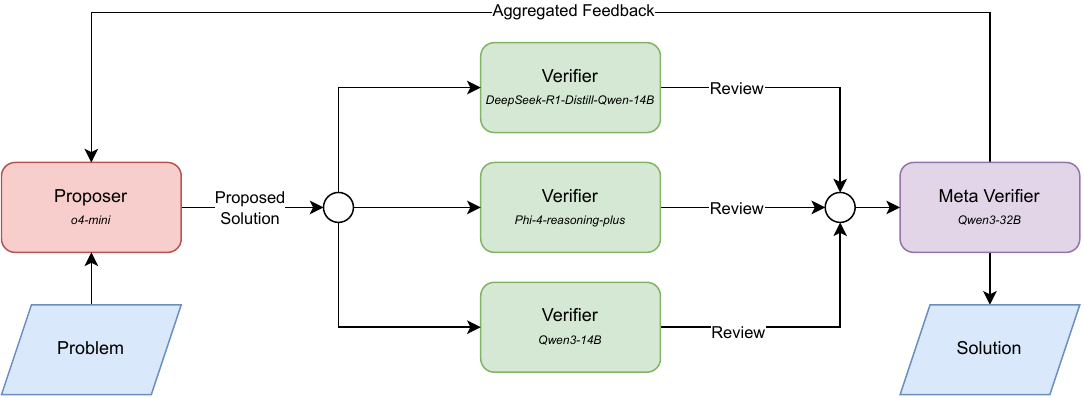}
    \setlength{\belowcaptionskip}{-3mm}
    \caption{An overview of the multi-agent review model. The model names are, of course, subject to shuffling.}
    \label{fig:mar-flowchart}
\end{figure*}

One of the techniques that we use to try to improve the performance of the LLMs is to have several smaller models review their proposed solutions. The subsections delineated below describe this technique in more detail. The method is illustrated in Figure~\ref{fig:mar-flowchart}.

\subsubsection{Proposer Module}
Given a set of physics problems \( \mathcal{Q} = \{ q_1, q_2, \ldots, q_N \} \), the proposer model \( \mathcal{P}_\theta \), instantiated as a powerful LLM (such as, OpenAI's o4-mini or Gemini 2.5 Pro), generates an initial set of solutions \( \mathcal{S}^{(0)} = \{ s_1^{(0)}, s_2^{(0)}, \ldots, s_N^{(0)} \} \), where each solution is sampled according to \( s_i^{(0)} \sim \mathcal{P}_\theta(\cdot \mid q_i) \). This zero-feedback stage constitutes the base output, from which all subsequent verification and refinement occur.

\subsubsection{Verifier Module}
\label{subsec:verifier}
To assess the quality of initial solutions generated by the proposer model, we employ a set of three large language models---Microsoft's Phi-4-reasoning-plus \citep{abdin2024phi}, Qwen3-14B \citep{yang2025qwen3}, and DeepSeek-R1 14B \citep{guo2025deepseek} —as independent verifiers. These models operate with fine-tuning and evaluate each proposed solution against the original problem statement.

Each verifier produces a structured evaluation comprising six scores, each ranging from 0 to 5, based on the manually engineered rubrics, inspired by \citet{docktor2016assessing}, outlined in Section \ref{sec:rubrics}.
A weighted average of these scores yields a final scalar score, with greater emphasis placed on formulation (0.25), numerical correctness (0.30), and logical consistency (0.25). The remaining weights are: completeness (0.10), validity of assumptions (0.05), and clarity (0.05). The aggregated score provides a quantitative measure of solution quality, while the individual component scores offer interpretable feedback, which is later utilized by the meta-verifier for cross-verifier analysis and refinement. Each verifier identifies and records perceived mistakes in the proposer’s solution, maintaining them in a structured mistake list.

\subsubsection{Meta-Verifier Module}
The meta-verifier module, Qwen3-32B by \citet{yang2025qwen3}, receives the test set questions, proposer solutions, and responses from three independent verifiers. The main task of the meta-verifier is to filter out \textit{irrelevant} or \textit{false} mistakes. Since the verifiers are independent, some mistakes flagged by one verifier may be irrelevant to the actual problem or inaccurate in the context of the solution. The meta-verifier assesses the validity of these mistakes by comparing them across all three verifiers and retains only those that are relevant and consistent. After filtering the mistakes, the meta-verifier aggregates the scores from all three verifiers into a final score using the weighted sum in Equation \ref{eq:eq1}.
\begin{multline}
\label{eq:eq1}
r_i^{(\operatorname{final})} = 0.5 \times r_i^{(\operatorname{Phi-4})} + \\
0.3 \times r_i^{(\operatorname{DeepSeek-R1})} + 
0.2 \times r_i^{(\operatorname{Qwen3-14B})}
\end{multline}

This final aggregated score reflects the overall quality of the proposer’s solution for each question \(q_i\). The weights are chosen based on the anecdotally observed quality and accuracy of each model's review responses. They are specifically determined based on a pilot study conducted on a randomly sampled set of 500 problems from our dataset. Under a zero-shot setting, the three verifiers' outputs are manually inspected for factual accuracy, reasoning depth, and alignment with the grading rubric. We find that Phi-4-reasoning-plus consistently provides the most precise and well-justified critiques, followed by DeepSeek-R1, and then Qwen3. The weights are set to reflect this observed performance hierarchy, with more gravitas accorded to the more reliable verifier. The meta-verifier’s output consists of:\\
\textbf{Aggregated Mistakes List:} A refined list of mistakes that have been validated across all 3 verifiers.\\
\textbf{Aggregated Score:} A weighted average of the individual scores from the 3 verifiers, reflecting the overall quality of the proposer’s solution. This refined approach helps improve the robustness and accuracy of the evaluation, guiding further refinement of the proposer’s solutions based on reliable and consistent feedback.

\subsection{Experimental Setup}
We use a total of six frontier LLMs to solve 1,962 physics problems from the test split of \physicseval. The generated solutions are the base solutions or \textit{proposed} solutions.

\subsubsection{Self-refining}
After receiving the proposed answer from the solver LLM, we provide its own solution and question again with the following additional metacognitive prompt:
\texttt{``You are a Physics Professor. Outline the physics principles of the given problem, and please check your own answers for any mistakes, then answer again."} The generated answer is henceforth considered the final answer.

\subsubsection{Single-Agent Review}
After receiving the proposed answer from the solver LLM, we get another LLM, namely Qwen3-32B, to analyze the question-answer pair and generate a list of probable mistakes. If any mistakes are found, we provide this list as additional information and the solver model's own solution to the solver model itself, and prompt it to solve the problem again.

\subsubsection{Multi-Agent Review}
We have the reviewer agents generate a list of mistakes for every pair of problem and solution. The meta-verifier agent then compiles these mistakes into a single list. These mistakes are then sent to the solver LLM along with its previous solution, and a new solution is requested, albeit only if the meta-reviewer finds any mistakes. A sample Multi-Agent Review conversation can be found in Appendix \ref{appendix:sample_convo}.











\subsection{Performance Evaluation and Metrics}
For a thorough evaluation of our inference-time techniques and comparison to baseline LLM performance, we use a comprehensive, multi-layered method. This approach breaks down the problem-solving process into key components essential for success in physics, enabling both detailed and overall assessment. Each solution generated by the different LLM configurations (baseline, w/ self-refinement, w/ single-agent verification, and w/ multi-agent verification) is evaluated against the ground-truth solution in \physicseval. These ground-truth solutions, derived from established textbook explanations and further elaborated by domain experts for clarity on step-by-step reasoning, serve as the definitive reference for correctness and methodology.

\subsection{Rubric Engineering for Evaluation}
\label{sec:rubrics}
The evaluation is conducted across six core qualitative and quantitative criteria (similar to the criteria used by the verifier module in multi-agent-verification as shown in Subsection~\ref{subsec:verifier}) denoted as \( M_k \) where \( k \in \{1, \dots, 6\} \). Each criterion is scored on a Likert scale from 1 to 5 (where 5 represents the highest quality):\\
\textbf{Mathematical Accuracy (\(S_\text{MA}\)):} Assesses the correctness of calculations, numerical answers, units, and appropriate presentation, strictly in comparison to the ground-truth answer.\\
\textbf{Logical Consistency (\(S_\text{LC}\)):} Evaluates the soundness of the step-by-step reasoning and its alignment with physics principles and the ground-truth solution's logic.\\
\textbf{Completeness (\(S_\text{C}\)):} Measures whether all parts of the problem, as scoped and addressed by the ground-truth solution, were fully addressed.\\
\textbf{Clarity and Coherence (\(S_\text{CC}\)):} Judges the clarity, conciseness, organization, and ease of understanding of the AI's explanation and use of terminology.\\
\textbf{Formulas and Principles (\(S_\text{FP}\)):} Determines if the correct physical formulas and principles were identified, stated, and applied appropriately by the AI, consistent with the problem's framing in the ground-truth.\\
\textbf{Assumptions Made (\(S_\text{A}\)):} Assesses whether the \textit{a priori} assumptions were clearly stated, justified, reasonable for the problem context, and did not contradict limitations identified by the ground-truth.\\
The score for a given solution \( j \) on metric \( k \) is denoted as \( s_{j,k} \in [1, 5] \).
\begin{table*}[t]
    \centering
    \resizebox{\textwidth}{!}{%
    \begin{tabular}{l cccc cccc cccc}
        \toprule
        & \multicolumn{4}{c}{\textbf{Easy (1-4)}} & \multicolumn{4}{c}{\textbf{Medium (5-7)}} & \multicolumn{4}{c}{\textbf{Hard (8-10)}} \\
        \cmidrule(lr){2-5} \cmidrule(lr){6-9} \cmidrule(lr){10-13}
        
        \textbf{Model} & Baseline & Self-Refine & Single-Agent & Multi-Agent & Baseline & Self-Refine & Single-Agent & Multi-Agent & Baseline & Self-Refine & Single-Agent & Multi-Agent \\
        \midrule
        
        DeepSeek-R1      & 90.6 & 92.7 & 93.3 & \textbf{94.1} & 80.8 & \textbf{84.4} & 83.1 & 83.4 & 72.9 & 73.7 & \textbf{74.7} & 72.7 \\
        
        Gemma 3 27B        & 86.9 & 85.7 & 86.4 & \textbf{87.6} & 55.8 & 56.5 & \textbf{59.4} & 59.1 & \textbf{41.5} & 40.4 & 39.1 & 40.6 \\
        
        Llama 4 Maverick  & 91.5 & 90.5 & 92.0 & \textbf{92.9} & \textbf{83.6} & 82.0 & 82.6 & 82.4 & 55.2 & 54.0 & \textbf{57.9} & 52.1 \\
        
        Phi-4-reasoning-plus  & 84.4 & 94.4 & 93.2 & \textbf{94.7} & 86.9 & 92.2 & 93.4 & \textbf{93.9} & 80.1 & 83.4 & 83.8 & \textbf{87.6} \\
        
        QwQ-32B         & 93.7 & 94.0 & 94.2 & \textbf{94.6} & 80.9 & 80.8 & 81.1 & \textbf{81.9} & 63.9 & 68.6 & 63.5 & \textbf{71.0} \\

        o4-mini        & 86.7 & 82.6 & 85.9 & \textbf{86.8} & 87.3 & 86.4 & 87.0 & \textbf{88.2} & 83.6 & 82.3 & 84.1 & \textbf{85.4} \\
        
        \midrule
        \textbf{Average} & 88.96  & 89.98 & 90.83 & \textbf{91.78} & 79.21 & 80.38 & 81.10 & \textbf{81.48} & 66.20 & 67.06 & 67.18 & \textbf{68.23} \\
        \bottomrule
    \end{tabular}
    } 
    \setlength{\belowcaptionskip}{-3mm}
    \caption{
        \textcolor{black}{Average PPS score on \physicsevals across frontier LLMs, stratified by problem difficulty and inference method.
        The "Average" row summarizes results. Higher scores are better. The best-performing method for each model and difficulty tier is highlighted in \textbf{bold}.}
    }
    \label{tab:pps-results}
\end{table*}
\subsection{Physics Proficiency Score (PPS)}
To capture overall problem-solving ability, we define a Physics Proficiency Score (PPS)—a weighted average of six key evaluation metrics. For a given solution \( j \), PPS is calculated as:
\begin{equation}
    \operatorname{PPS}_j = \sum_{k=1}^{6} w_k \cdot s_{j,k}
    \label{eq:pps}
\end{equation}
where, \( s_{j,k} \) denotes the score for solution \( j \) on metric \( k \), and \( w_k \) is the weight assigned to metric \( k \). The weights reflect the importance of each metric in solving physics problems accurately. These weights are shown in Table~\ref{tab:tab1}. To ground these weights in practical educational assessment, we consulted with two experienced high school physics teachers and one university-level entrance examination grader. The goal was to capture their professional judgment on the relative importance of each criterion when evaluating a student's solution. The high weight on Mathematical Accuracy and Logical Consistency reflects the consensus that a correct answer derived from sound reasoning is the paramount goal, while Clarity, though important, is often secondary in formal grading. This process of \textit{expert elicitation} ensures our scoring aligns with human assessment priorities.
\begin{table}[t]
    \centering
    \resizebox{0.85\linewidth}{!}{%
    \begin{tabular}{lc}
        \toprule
        Metric Component & Weight (\(w_k\)) \\
        \midrule
        Mathematical Accuracy (\(w_\text{MA}\)) & 0.30 \\
        Logical Consistency (\(w_\text{LC}\)) & 0.25 \\
        Formulas and Principles (\(w_\text{FP}\)) & 0.20 \\
        Completeness (\(w_\text{C}\)) & 0.10 \\
        Assumptions Made (\(w_\text{A}\)) & 0.10 \\
        Clarity and Coherence (\(w_\text{CC}\)) & 0.05 \\
        \midrule
        \textbf{Total} & \textbf{1.00} \\
        \bottomrule
    \end{tabular}
    }
    \setlength{\belowcaptionskip}{-5mm}
    \caption{Weights for Physics Proficiency Score (PPS) sub-metrics}
    \label{tab:tab1}
\end{table}
This scoring system emphasizes correctness---especially math, logic, and the proper use of physical principles. Completeness and sound assumptions also matter, while clarity is valued but given less weight. The final PPS is a value that is normalized to be $\in [0,100]$.

\subsection{Evaluation Approach Justification}
The core philosophy of our rubric (see \autoref{tab:tab1}) is adapted from the Minnesota Assessment of Problem Solving (MAPS) framework \cite{docktor2016assessing}, a research-validated tool for evaluating student-written solutions in introductory physics courses. MAPS was designed to assess the quality of problem-solving processes rather than just final answers, drawing on extensive physics education research to identify key cognitive steps that distinguish effective reasoning. Its underlying rationale is that strong problem-solving involves a structured, logical progression that accurately interprets the physical scenario, selects and applies appropriate principles, performs correct mathematics, and maintains coherence. In our adaptation for \physicseval, we aligned our rubric categories with MAPS's five core processes:
    Useful Problem Description (organizing given information and visualizing the scenario),
    Physics Approach (selecting relevant concepts and principles),
    Specific Application of Physics (applying those principles to the situation),
    Mathematical Procedures (executing calculations correctly), and
    Logical Progression (ensuring overall coherence without gaps or contradictions).

This fidelity-focused emphasis ensures that LLM-generated solutions are scored not only for superficial correctness but for alignment with the ground truth's expert-derived interpretation (\textit{e.g.}, correctly framing the problem's physics), method (\textit{e.g.}, step-by-step application), and result (\textit{e.g.}, final answer) with minimal subjectivity. For instance, a solution that reaches the right numerical answer via a fallacious principle would score low on "Formulas and Principles" fidelity, reflecting real-world grading practices where process matters as much as product.

\begin{table*}[t]
    \centering
    \resizebox{0.9\textwidth}{!}{%
    \begin{tabular}{ll rrrc} 
        \toprule
        \textbf{Comparison} & \textbf{Metric} & \textbf{Weight} & \textbf{$t$-statistic} & \textbf{$p$-value} & \textbf{Significant $(p < 0.05)$?} \\
        \midrule
Single Agent \textit{vs.} Baseline & Overall PPS & 1.00 & -0.01 & 0.996 & False \\
Single Agent \textit{vs.} Baseline & Mathematical Accuracy & 0.30 & 0.89 & 0.373 & False \\
Single Agent \textit{vs.} Baseline & Logical Consistency & 0.25 & 0.35 & 0.729 & False \\
Single Agent \textit{vs.} Baseline & Completeness & 0.10 & -2.36 & 0.0182 & \cellcolor{highlightgreen}True \\
Single Agent \textit{vs.} Baseline & Clarity And Coherence & 0.05 & -3.31 & 0.000965 & \cellcolor{highlightgreen}True \\
Single Agent \textit{vs.} Baseline & Formulas Principles & 0.20 & 0.29 & 0.769 & False \\
Single Agent \textit{vs.} Baseline & Assumptions Made & 0.10 & -1.23 & 0.220 & False \\
\midrule 
Multi-Agent Review vs. Baseline & Overall PPS & 1.00 & 2.05 & 0.0405 & \cellcolor{highlightgreen}True \\
Multi-Agent Review vs. Baseline & Mathematical Accuracy & 0.30 & 2.77 & 0.00570 & \cellcolor{highlightgreen}True \\
Multi-Agent Review vs. Baseline & Logical Consistency & 0.25 & 1.08 & 0.278 & False \\
Multi-Agent Review vs. Baseline & Completeness & 0.10 & -1.44 & 0.150 & False \\
Multi-Agent Review vs. Baseline & Clarity And Coherence & 0.05 & -1.79 & 0.0739 & False \\
Multi-Agent Review vs. Baseline & Formulas Principles & 0.20 & 2.50 & 0.0126 & \cellcolor{highlightgreen}True \\
Multi-Agent Review vs. Baseline & Assumptions Made & 0.10 & 1.50 & 0.134 & False \\
        \bottomrule
    \end{tabular}
    } 
    \setlength{\belowcaptionskip}{-3mm}
    \caption{Statistical significance of performance improvements for o4-mini. Results with a $p$-value $< 0.05$ are considered statistically significant.}
    \label{tab:sig_test}
\end{table*}
\subsection{Evaluation Process}

\label{sec:evaluation_process}
Solutions were evaluated using Gemini 2.5 Pro. The LLM was provided with a detailed scoring rubric, the ground-truth solution, and the AI-generated solution for each problem. The evaluation prompt provided to Gemini 2.5 Pro
was designed to ensure strict comparison against the ground truth. This prompt guided the assignment of scores (1--5 for sub-metrics). We select Gemini 2.5 Pro as the evaluator because independent leaderboards, \textit{e.g.}, LMSys Chatbot Arena (LMArena)\footnote{\url{https://lmarena.ai/leaderboard}}, consistently rank it as the best model for text generation, complex instruction following, and nuanced judgment. This makes it well-suited for applying our detailed, rubric-based scoring system consistently and accurately. To prevent any implicit favoritism toward its own generation style, we designed the evaluation prompt to strictly compare generated solutions against the ground-truth (human-derived) solutions using objective, predefined criteria (\textit{e.g.}, mathematical accuracy, logical consistency). The prompt explicitly instructs the model to ignore stylistic deviations (see Appendix \ref{appendix:eval_prompt2}) and focus on alignment with the ground truth's content and methodology, reducing the risk of bias. We also performed empirical checks on a subset of evaluations manually and ensured that the model maintained high fidelity to the rubrics without apparent bias.

\subsection{Statistical Analysis}
The collected scores (both for individual metrics \(s_{j,k}\) and the aggregated \(\operatorname{PPS}_j\)) are analyzed using descriptive statistics (mean, median, standard deviation) for each metric and each LLM configuration. 
Performance is also analyzed across different physics categories and problem complexity levels (proxied by solution length). Table \ref{tab:pps-results} portrays the results yielded by all the pertinent models when evaluated on our \physicsevals benchmark. 

\section{Results Analysis}
\subsection{Key Findings}
The evaluation results on the \physicsevals benchmark, as delineated in Table \ref{tab:pps-results}, reveal several key insights into the effectiveness of different agent-based inference methods.

\paragraph{Agent-Based Critique Consistently Improves Performance:} A primary observation across all models and difficulty tiers is the general trend of performance improvement with more sophisticated methods. On average, the vanilla Baseline scores are the lowest, followed by incremental gains from Self-Refine, Single-Agent critique, and finally the Multi-Agent framework, which achieves the highest average scores across Easy (91.78), Medium (81.48), and Hard (68.23) problems. This demonstrates that external critique is a more reliable enhancement strategy than simple self-correction.

\paragraph{Gains from Multi-Agent System Amplify with Problem Difficulty:} While all methods experience a performance drop as problems become harder, the advantage of the Multi-Agent system over the Baseline becomes more pronounced. For instance, on Hard problems, Phi-4-reasoning-plus gains 7.5 points with the Multi-Agent method over its baseline, and QwQ-32B gains 7.1 points. This suggests that the collaborative verification and feedback process is particularly valuable for tackling complex reasoning tasks where a single model is more likely to fail.
\begin{figure*}[t]
    \centering
    \includegraphics[width=\linewidth]{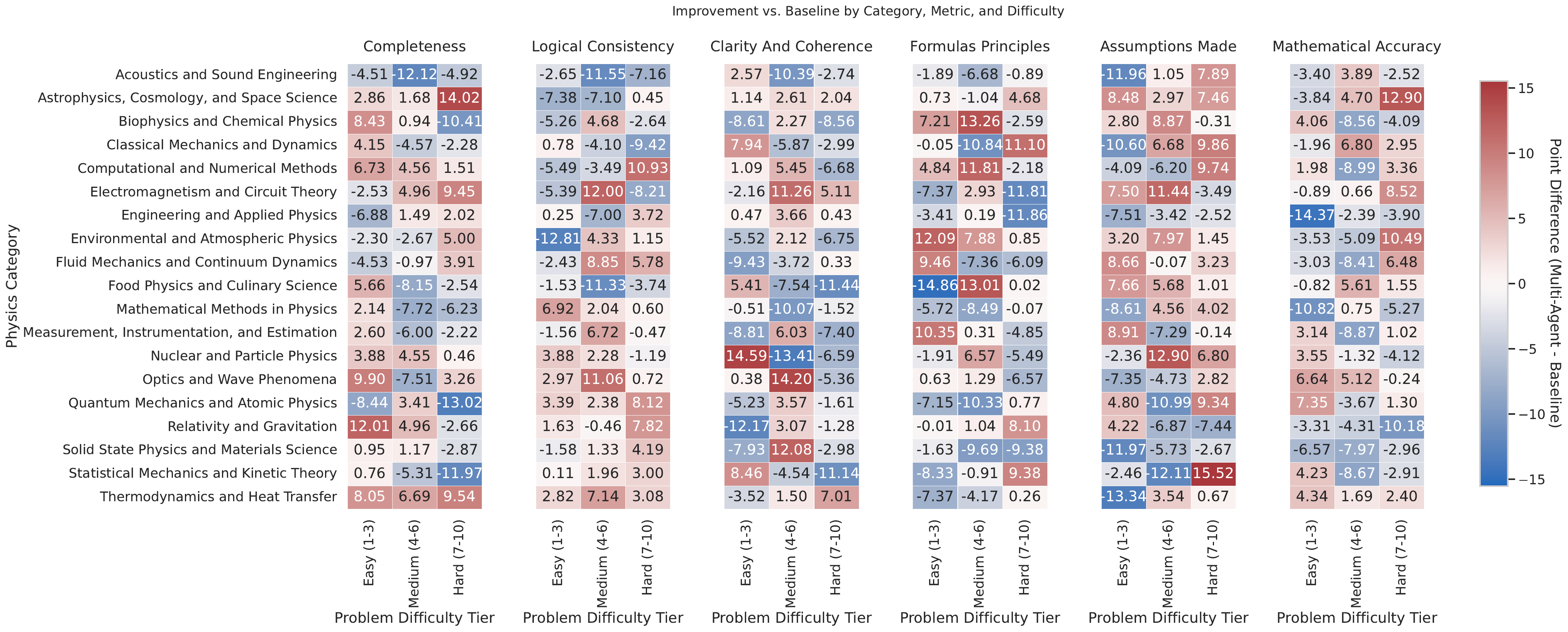}
    \setlength{\belowcaptionskip}{-3mm}
    \caption{Category-specific impact of the Multi-Agent Review framework across all scoring rubrics for o4-mini.}
    \label{fig:categorywise_performance}
\end{figure*}

The effectiveness of each method is highly dependent on the underlying model. Phi-4-reasoning-plus stands out as the top-performing model, achieving the highest scores in 7 of the 12 categories. It benefits significantly from all advanced methods, showcasing a strong capacity for both self-correction and assimilating external feedback. o4-mini also proves to be a robust model, establishing the highest baseline for Medium and Hard problems and showing consistent improvement with agent-based methods.

A notable and counterintuitive finding is that the self-refine process can harm performance. For models like Gemma 3 27B \cite{gemma-report} and Llama 4 Maverick \cite{meta2025llama4maverick}, self-refinement consistently resulted in lower scores compared to their own baseline across all difficulty levels. This indicates that without external guidance, a model's attempt to \textit{"double-check"} its work can introduce new errors or reinforce incorrect initial assumptions, making it an unreliable strategy for certain architectures.

\subsection{Results Analysis and Discussion}
\label{sec:results}
In order to delve deeper into the performance analysis on \physicseval, we keep OpenAI's o4-mini under the limelight. A detailed examination of the model's performance on \physicsevals reveals a nuanced relationship between the model's inherent capabilities and the efficacy of agent-based refinement methods. The model exhibits a formidable baseline proficiency, with an initial PPS of 85.88, underscored by particularly high scores in Clarity and Coherence (4.76) and Formulas/Principles (4.59). This suggests the model's native strength lies in articulating solutions clearly and correctly identifying the underlying physics. However, the Self-Refined approach proved detrimental, leading to a notable decline in the overall PPS to 84.58. A closer look at the sub-metrics uncovers that while Mathematical Accuracy saw a slight improvement (4.17 to 4.22), this came at a significant cost to Completeness (4.56 to 4.40) and Clarity (4.76 to 4.54). This paradox indicates that without external guidance, the model's attempt to self-correct can disrupt the solution's logical structure and narrative coherence. In contrast, while the Single Agent method offered negligible improvement, the Multi-Agent Review emerges as the only method to yield a definitive performance enhancement, elevating the PPS to 86.84. This gain is not merely marginal but is driven by targeted improvements in the model's weakest areas—Mathematical Accuracy (4.24), Logical Consistency (4.56), and Assumptions Made (4.43)—while largely preserving its strong baseline clarity. This finding powerfully suggests that for a highly capable model like o4-mini, further advancement is not achieved through simple self-correction but through a robust, consensus-driven verification process that can surgically address specific logical and computational flaws without compromising the solution's overall quality.




\subsubsection{Category-Specific Impact of the Multi-Agent Framework}
The impact of Multi-Agent Review on o4-mini across different categories can be seen in Figure \ref{fig:categorywise_performance}. It reveals that the impact of multi-agent review on o4-mini’s performance is highly category- and rubric-dependent, with both substantial gains and notable degradations across physics domains and difficulty tiers. Noteworthy improvements are observed in categories such as Quantum Mechanics and Atomic Physics, Relativity and Gravitation, and Thermodynamics and Heat Transfer, particularly for hard problems and rubrics like Completeness and Formulas/Principles. Medium-difficulty problems in areas like Food Physics and Culinary Science, Optics, and Wave Phenomena also see dramatic gains in Clarity and Coherence. However, the benefits are not universal: some categories, including Acoustics and Sound Engineering and Engineering and Applied Physics, experience negative or inconsistent changes, especially for easier problems, and certain rubrics, such as Assumptions Made and Mathematical Accuracy, even show degradations in select domains. These results highlight that while multi-agent review can substantially enhance performance in specific contexts—especially for complex, technical, or computationally intensive tasks—it may also introduce confusion or diminish quality in others, underscoring the need for more adaptive and context-aware review strategies. Additional details can be seen in Figure \ref{fig:mar_pps_model_category} in Appendix \ref{sec:further_info}.


\subsubsection{Statistical Significance of Performance Changes}

To evaluate the effectiveness of different inference-time techniques, we perform paired $t$-tests comparing each method against the o4-mini baseline, using a significance threshold of $p < 0.05$. The Single Agent method fails to produce any statistically significant improvement in Overall PPS $(p = 0.996)$, and instead shows significant degradations in Completeness $(p = 0.0182)$ and Clarity and Coherence $(p = 0.000965)$. This suggests that introducing a single external reviewer may negatively impact the structural and narrative quality of the generated solutions, potentially by injecting inconsistent or insufficient feedback. On the other hand, the Multi-Agent Review approach demonstrated statistically significant improvements in Overall PPS $(p = 0.0405)$, Mathematical Accuracy $(p = 0.0057)$, and Formulas Principles $(p = 0.0126)$. These gains indicate that collaborative critique across multiple agents can more effectively guide the model toward better mathematical correctness and principled reasoning. Although improvements in other dimensions, such as Logical Consistency and Completeness, do not reach significance, the overall results imply the advantage of multi-agent systems in enhancing both the accuracy and interpretability of the model's solutions.

\section{Conclusion}
Our comprehensive evaluation of frontier LLMs on the newly introduced \physicsevals benchmark demonstrates both the promise and the current limitations of LLMs in the domain of physics problem-solving. While baseline model performance is already strong for many categories, our experiments reveal that agentic inference-time techniques---particularly multi-agent verification---can yield substantial improvements, especially for challenging problems and in technical subfields where initial model outputs are less reliable. However, these gains are not uniform across all categories and rubrics, highlighting the nuanced and context-dependent nature of collaborative critique. Our findings underscore the importance of adaptive, category-aware strategies for further advancing LLM capabilities in scientific reasoning. We hope that \physicsevals will serve as a valuable resource for the research community.

\section*{Limitations}
One main limitation of our approach is the high computational cost. The multi-agent review method runs several models for each question, which increases processing time and resource use. This makes it less practical for real-time or low-resource settings. Moreover, our dataset, while large and varied, didn't undergo full manual checking. The human-derived solutions were stylistically refined using an LLM (Gemini 2.5 Pro), and while this process was designed to preserve the original logic, only a small sample was manually reviewed to confirm this. This means there could be mistakes in the data that affect model performance and evaluation. Due to the elaborate and descriptive nature of the ground truth, the evaluation of the solutions is largely LLM-based. Lastly, while our methods work well for physics problems, they may not transfer easily to other STEM areas without changes or fine-tuning.
\section*{Ethics Statement}
This work involves the use of publicly available large language models and does not include any human subjects, private data, or personally identifiable information. All physics problems are collected from publicly available sources, and care is taken to ensure that no copyrighted or proprietary content is used without proper attribution. The dataset partially relies on LLM-generated content for elaboration, which may introduce unintended biases or inaccuracies. We acknowledge this limitation and emphasize the importance of responsible use and validation in downstream applications. The proposed multi-agent inference techniques aim to improve performance without retraining, allowing broader access to high-performing systems. However, we recognize the increased computational cost associated with such approaches and encourage mindful deployment, especially in energy-sensitive or resource-constrained environments. We support transparency and reproducibility and plan to release the dataset and code where licensing permits.

\section*{Acknowledgements}
We convey our heartfelt gratitude to the anonymous reviewers for their constructive criticisms and insightful feedback, which were conducive to the improvement of the research work
outlined in this paper. We also appreciate the Systems and Software Lab (SSL) of the Islamic University of Technology (IUT) for the generous provision of computing resources during the course of this project. Syed Rifat Raiyan, in particular, wants to thank his parents, Syed Sirajul Islam and Kazi
Shahana Begum, for everything.

\bibliography{custom}
\bibliographystyle{acl_natbib}
\clearpage
\appendix
\onecolumn

\section{Evaluation Prompt for Gemini 2.5 Pro}
\label{appendix:eval_prompt2}
\begin{tcolorbox}[colback=gray!5, colframe=black, title={Prompt}]
You are an expert physics problem evaluator. Your task is to meticulously and STRICTLY compare an AI-generated solution to a manual, ground-truth solution for a given physics problem. The Ground Truth Solution is considered the definitive correct answer and approach for the given problem statement. Deviations by the AI-Generated Solution from the Ground Truth, especially in terms of method, assumptions, interpretation of given data, or parts deemed unsolvable by the Ground Truth, MUST be penalized appropriately according to the guidelines below. 

Evaluate the AI-generated solution based on the following categories and scoring guidelines. Provide your evaluation STRICTLY as a JSON object. 

Evaluation Categories and Scoring Guidelines: 

1.  mathematical\_accuracy: (Score 1-5) How correct are the AI's calculations, numerical answers, and units *when compared to the problem defined by the Ground Truth*? 

  - 5: All calculations, numerical results, and units are perfectly correct and appropriately presented, AND align with the Ground Truth's final answers if the same method is used, OR are verifiably correct if a different valid method is used. 

  - 4: Minor calculation error in the AI solution, or an incorrect/missing unit, but the AI's underlying mathematical method (if aligned with GT or a valid alternative) is sound. 

  - 3: Several minor errors in the AI solution, or one significant calculation error that impacts the AI's result. Units might be inconsistently handled. 

  - 2: Major calculation errors or fundamental misunderstandings of mathematical operations in the AI solution. If the AI solution uses different input data values than implied by the Ground Truth (e.g., different length, mass), leading to numerically different answers, score 2 here even if its internal math is correct for its chosen data, because it's not solving the *Ground Truth's* problem. 

  - 1: Almost all calculations in the AI solution are incorrect, non-sensical, or missing. The AI uses drastically different input data leading to completely irrelevant numerical results for the Ground Truth problem. 

2.  logical\_consistency: (Score 1-5) Does the AI solution follow a logical step-by-step progression? Is the AI's reasoning sound and aligned with physics principles, *ideally mirroring or compatibly extending the Ground Truth's logic*? 

  - 5: The AI solution flows perfectly. Each step logically follows from the previous one. The reasoning is impeccable and aligns well with the Ground Truth's approach or is a valid alternative. 

  - 4: AI solution is mostly logical and well-reasoned. Perhaps one step is slightly unclear or its justification is weak, but it doesn't break the overall logic or significantly deviate from a valid path. 

  - 3: Some logical gaps, inconsistencies, or steps in the AI solution that don't clearly follow, making the solution harder to follow or verify, or deviating from the core logic of the Ground Truth without clear justification. 

  - 2: Significant logical flaws in the AI solution. Steps are out of order, reasoning is poor or contradictory to established physics or the Ground Truth's interpretation. 

  - 1: The AI solution is illogical, incoherent, or internally contradictory. 

3.  completeness: (Score 1-5) Does the AI-generated solution address all parts of the problem *as understood and scoped by the Ground Truth*? 

  - 5: All parts of the problem (including sub-questions, if any), as addressed or implied as solvable by the Ground Truth, are fully addressed and answered by the AI. 

  - 4: A minor aspect of the problem (as per GT) is overlooked by the AI, or one sub-question is not fully answered or is missing. 
\end{tcolorbox}
\begin{tcolorbox}[colback=gray!5, colframe=black]
  - 3: A significant part of the problem (as per GT) is ignored or left unanswered by the AI. If the Ground Truth indicates a part of the problem is unsolvable with given data, but the AI attempts to solve it by making significant unstated/unwarranted assumptions, this is a flaw in understanding problem scope; score 3 or lower. 

  - 2: Only a small portion of the problem (as per GT) is addressed by the AI; major components are missing. 

  - 1: The problem is largely unaddressed by the AI, or the AI solution is off-topic relative to the Ground Truth. 

4.  clarity\_and\_coherence: (Score 1-5) Is the AI's explanation clear, concise, and easy to understand? 

  - 5: The AI explanation is exceptionally clear, concise, well-structured, and very easy to understand. Excellent use of language and terminology. 

  - 4: The AI explanation is clear and generally easy to understand, with minor areas for improvement in conciseness, structure, or flow. 

  - 3: The AI explanation is generally understandable but may be verbose, unclear in parts, poorly organized, or contain jargon without adequate explanation. 

  - 2: The AI explanation is difficult to understand due to ambiguity, poor writing, or convoluted structure. 

  - 1: The AI explanation is incomprehensible, extremely poorly written, or nonsensical. 

5.  formulas\_principles: (Score 1-5) Are correct physical formulas and principles identified and applied correctly by the AI, *and are they appropriate for the problem as framed by the Ground Truth*? 

  - 5: All necessary physical formulas and principles are correctly identified, stated, and applied appropriately by the AI, consistent with the Ground Truth's approach or a valid, equally rigorous alternative. 

  - 4: Mostly correct formulas/principles used by AI. Perhaps a minor error in recalling a formula, or a slight misapplication of a correct principle that doesn't fundamentally alter the solution path compared to GT. 

  - 3: Some incorrect formulas/principles are used by AI, or correct ones are applied incorrectly in a significant way. Or, the AI uses a principle that oversimplifies the problem compared to the level of detail expected by the Ground Truth. 

  - 2: Major errors in formula/principle selection or application by AI. Fundamental physics concepts are misunderstood by the AI. 

  - 1: Completely inappropriate formulas/principles are used by AI, or relevant physics is entirely ignored. 

6.  assumptions\_made: (Score 1-5) Are AI assumptions (explicit or implicit) explicit, justified, and reasonable *especially when compared to the Ground Truth's scope and stated/implied assumptions*? 

  - 5: All necessary assumptions made by the AI are explicitly stated, well-justified, and perfectly reasonable for the problem context, AND do not contradict or bypass limitations identified by the Ground Truth. 

  - 4: Most necessary assumptions made by the AI are stated and reasonable; some minor ones might be implicit but obvious, or lack full justification but are acceptable and align with GT. 

  - 3: Some key assumptions in the AI solution are missing, not clearly stated, or questionable in reasonableness. Or, the AI makes assumptions that simplify the problem in a way the Ground Truth does not. 

  - 2: Major unreasonable assumptions are made by the AI, or critical assumptions are not stated, leading to an incorrect or flawed solution path. This includes assumptions that allow solving parts the Ground Truth indicates are unsolvable with the given data. 
\end{tcolorbox}

\begin{tcolorbox}[colback=gray!5, colframe=black]
      - 1: Assumptions in the AI solution are entirely inappropriate, absent when clearly needed, or lead to a trivialization/misrepresentation of the problem as defined by the Ground Truth. 

7.  overall\_correctness: (Score 0-10) How correct and sound is the AI's approach and final answer(s) overall, *primarily judged by its fidelity to the Ground Truth's interpretation, method, and result for the specific problem*? 

  - 10: Perfect solution. The AI's method, reasoning, data interpretation, assumptions, and final answer(s) align flawlessly or are an equally valid and rigorous path to the Ground Truth. 

  - 8-9: Excellent solution. Fundamentally correct with very minor, inconsequential flaws or slight stylistic deviations from the Ground Truth, but arrives at the same essential understanding and results. 

  - 6-7: Good solution. Generally correct approach by the AI, and largely correct answer(s), but with some noticeable errors, omissions, or areas for improvement when compared to the Ground Truth. The AI might use a valid but less ideal method. 

  - 4-5: Partially correct. The AI demonstrates some understanding but contains significant flaws in reasoning, calculation, choice of principles, or makes unwarranted assumptions that lead it away from the Ground Truth's solution. This score is appropriate if the AI solves a simplified version of the problem or misses key constraints implied by the Ground Truth. 

  - 2-3: Mostly incorrect. The AI shows fundamental misunderstandings of the problem or physics principles as defined by the Ground Truth. A solution that uses 

*different fundamental input data* than the Ground Truth CANNOT be rated higher than 3, even if its internal logic is sound for its chosen data. 

  - 0-1: Completely incorrect, irrelevant, or no meaningful attempt made by the AI to solve the problem as presented and solved by the Ground Truth. 

Problem ID: {problem\_id} 

Ground Truth Solution (this is the reference correct solution): 

{ground\_truth} 

Elaborated Solution Steps (this explains the Ground Truth): 

Elaborated Solution Steps (Manual): {elaborated\_solution} 

AI-Generated Solution to Evaluate (compare this against the Ground Truth and Elaborated Solution): {ai\_solution} 

Provide your evaluation STRICTLY as a JSON object with the problem\_id and scores for each category listed above. Your entire response should be ONLY the JSON object, starting with \texttt{\{\{} and ending with \texttt{\}\}}. 

\begin{lstlisting}[basicstyle=\ttfamily, breaklines=true]
Example JSON format: 
{{ 
    "problem_id": "{problem_id}", 
    "mathematical_accuracy": <score_1_to_5>, 
    "logical_consistency": <score_1_to_5>, 
    "completeness": <score_1_to_5>, 
    "clarity_and_coherence": <score_1_to_5>, 
    "formulas_principles": <score_1_to_5>, 
    "assumptions_made": <score_1_to_5>, 
    "overall_correctness": <score_0_to_10> 
}} 
\end{lstlisting}
\end{tcolorbox}
\twocolumn
\section{Sample Conversation with o4-mini (Multi-Agent Review)}
\label{appendix:sample_convo}
Here, we have done multi agent review on a proposed solution. The list of mistakes generated by the review process is fed back to the proposer.

\textbf{User:}
You are an expert on Physics. You solve problems step by step while maintaining logical consistency. Solve the following Physics problem: A toroidal inductor with an inductance of 9.0 mH encloses a volume of 0.0200 m$^3$. If the average energy density in the toroid is 70.0 J/m$^3$, what is the current through the inductor?
Finally, write the final answers in brief. Make sure you write all equations in \LaTeX.

\textbf{Assistant (o4-mini - Initial Solution):}
Here is the step-by-step solution:
1. Energy stored in an inductor
   $U = \frac{1}{2} L I^2$
2. Relation between total energy and energy density
   $U = u V$
   where $u$ = energy density, $V$ = volume
3. Equate the two expressions for $U$:
   $\frac{1}{2} L I^2 = u V$
4. Solve for the current $I$:
   $I = \sqrt{\frac{2 u V}{L}}$
5. Substitute the given values:
   $L = 9.0 \times 10^{-3} \text{ H}$
   $u = 70.0 \text{ J/m}^3$
   $V = 0.0200 \text{ m}^3$
   $I = \sqrt{\frac{2 \cdot 70.0 \cdot 0.0200}{9.0 \times 10^{-3}}} = \sqrt{\frac{2.80}{9.0 \times 10^{-3}}} = \sqrt{311.11} \approx 17.6 \text{ A}$

Final Answer (brief):
$I = \sqrt{\frac{2uV}{L}} \approx 17.6 \text{ A}$

\textbf{User (Feedback from Meta-Verifier based on Reviewers):}
I have some feedback. The solution assumes that the average energy density is uniformly distributed throughout the entire volume of the toroid. In reality, for a toroid, the magnetic field varies inversely with the distance from the center, leading to non-uniform energy density. The formula used assumes uniform energy density, which is an approximation. This assumption might not be accurate if the field varies significantly within the toroid. After taking this into account, please generate the solution once again. Remember to write all equations in \LaTeX.

\textit{[o4-mini would then generate a revised solution, ideally addressing the feedback.]}
\begin{table*}[ht]
\centering
\small  
\resizebox{\linewidth}{!}{%
\begin{tabular}{@{}p{5cm}p{8cm}@{}}
\toprule
\textbf{Field Name} & \textbf{Strength / Purpose} \\
\midrule
\texttt{Problem\_ID} & Enables consistent referencing, indexing, and analysis across models and experiments. \\
\texttt{problem} & Maintains fidelity to real-world phrasing typically encountered in education or exams. \\
\texttt{simplified\_problem\_statement} & Helps models/annotators focus on core reasoning, improving interpretability. \\
\texttt{category} & Enables domain-wise evaluation and curriculum design. \\
\texttt{soft\_labels} & Supports nuanced supervision and better error analysis. \\
\texttt{elaborated\_solution\_steps} & Crucial for explainability and multi-step reasoning evaluation. \\
\texttt{alternative\_solutions} & Promotes robustness and exposure to diverse reasoning. \\
\texttt{problem\_difficulty} & Enables benchmarking and curriculum learning by difficulty. \\
\texttt{final\_answers\_in\_brief} & Useful for accuracy checks and extractive answer training. \\
\texttt{steps} & Helps with curriculum learning and complexity analysis. \\
\bottomrule
\end{tabular}
}
\caption{Description of fields in the physics dataset and their purposes}
\label{tab:dataset_fields}
\end{table*}
\section{\physicseval: Additional Details}
\label{appendix:dataset_desc}
\label{app:dataset}
\subsection{Construction}
To enable large-scale evaluation and training of reasoning-capable language models in physics, we curate a comprehensive dataset of \textbf{19,609 annotated problems}, sourced from 20 different authoritative physics textbooks and verified educational websites.

The dataset spans 19 different categories, including \textit{Mechanics, Thermodynamics, Electromagnetism, Waves, Optics, Relativity}, and \textit{Quantum Physics}.

Each problem is processed through the following pipeline:

\begin{itemize}
    \item \textbf{Data Cleaning:} Raw content is cleaned to remove noise and inconsistencies.
    \item \textbf{\LaTeX  Annotation:} All equations are converted into \LaTeX  for structured mathematical representation.
    \item \textbf{Step-Wise Elaboration:} Using Gemini 2.5 Pro in ``Think'' mode, solutions are decomposed into logically coherent steps to enhance interpretability for LLMs. The pre-existing, human-authored solution steps were processed through Gemini 2.5 Pro. The model’s function was not to `generate' solutions, but to act as a sophisticated tool for stylistic refinement. Its role was strictly limited to improving textual clarity, correcting grammar and typographical errors, and standardizing the syntactic structure of the explanations. The core logic, mathematical formulations, and problem-solving path remained unchanged from the human-originated content.
    \item \textbf{Metadata Tagging:} Each problem is annotated with topic category, difficulty level, and key physical principles.
\end{itemize}

\subsubsection{Train-Test Split}
We apply a 90:10 split, resulting in \textbf{17,647 training} and \textbf{1,962 test samples}, supporting generalization across diverse reasoning tasks.

\subsubsection{Human-Annotated Difficulty}
The foundational difficulty labels (Easy, Medium, Hard) were inherited directly from the source material. When scraping problems from physics forums and educational websites, we preserved the difficulty ratings assigned by the original posters or instructors. These labels embody a human expert's judgment of the conceptual and reasoning demands of the problem.

\subsubsection{Solution Length \textit{vs.} Difficulty}
The number of steps in the elaborated solution is also stored. Some alternative solution methods are also suggested. Although length and complexity are not \textit{perfectly} synonymous, in the context of physics problem-solving, there is a demonstrable correlation. Longer solutions typically indicate problems that require more steps, more complex formulas, additional conceptual justification, or the synthesis of multiple physical principles. As shown in \autoref{tab:difficulty_metrics} (extracted from our dataset analysis), we observe a mostly monotonic relationship between annotated difficulty level and average solution length (measured in tokens). This trend supports our use of solution length as a pragmatic proxy for problem complexity.

\begin{table}[h!]
\centering
\small
\resizebox{\columnwidth}{!}{%
\begin{tabular}{c c c c}
\toprule
\textbf{Difficulty} & \textbf{Count} & \multicolumn{2}{c}{\textbf{Metrics}} \\
\cline{3-4}
 &  & \shortstack{\textbf{Avg.}\\\textbf{Solution}\\\textbf{Steps}} & \shortstack{\textbf{Avg.}\\\textbf{Solution}\\\textbf{Tokens}} \\
\midrule
1  & 47   & 3.383 & 2050.787 \\
2  & 489  & 3.589 & 2303.133 \\
3  & 1687 & 3.528 & 2529.343 \\
4  & 1450 & 3.675 & 2809.422 \\
5  & 2797 & 3.616 & 3034.966 \\
6  & 6718 & 3.852 & 3640.053 \\
7  & 5465 & 4.126 & 4788.111 \\
8  & 872  & 4.594 & 6523.017 \\
9  & 81   & 4.951 & 9209.000 \\
10 & 3    & 4.858 & 9201.010 \\
\bottomrule
\end{tabular}%
}
\caption{Solution metrics grouped by difficulty.}
\label{tab:difficulty_metrics}
\end{table}
This tessellates well with prior work, such as in MATH \cite{hendrycks2measuring} or GSM8K \cite{cobbe2021gsm8k}, where solution length serves as a computational indicator of reasoning depth. We acknowledge that length isn't perfect (\textit{e.g.}, a concise but conceptually dense problem might be underrated), so in future iterations, we could incorporate additional metrics like the number of distinct physical principles invoked or graph-based reasoning complexity.

\subsection{Data Model}
The dataset has the following fields:
\begin{itemize}
    \item \texttt{Problem\_ID}: Unique identifier for the problem instance
    \item \texttt{problem}: Original, full problem text from source material
    \item \texttt{simplified\_problem\_statement}: Paraphrased version, stripped of complexity
    \item \texttt{category}: Topical category (\textit{e.g.}, Mechanics, Optics)
    \item \texttt{soft\_labels}: Tags like numerical, conceptual, multi-step, diagram
    \item \texttt{elaborated\_solution\_steps}: Step-by-step reasoning to the correct answer
    \item \texttt{alternative\_solutions}: Different valid solution methods
    \item \texttt{problem\_difficulty}: Difficulty rating (1--10)
    \item \texttt{final\_answers\_in\_brief}: Final answer(s) only, no reasoning
    \item \texttt{steps}: Number of steps in main solution
    \item \texttt{source}: The source of the problem
\end{itemize}

\subsection{Dataset Sources}
\label{app:books}
\begin{itemize}
\item \textbf{21st Century Astronomy: Stars and Galaxies, 4th Edition} - by Laura Kay, Stacy Palen, Brad Smith, and George Blumenthal \cite{kay201621st}
\item \textbf{A Complete Resource Book for JEE Main 2018: Physics} - by Sanjeev Kumar
\item \textbf{Physics: Principles with Applications, 7th Edition} - by Douglas C. Giancoli \cite{giancoli2005physics}
\item \textbf{Physics for Scientists and Engineers: A Strategic Approach with Modern Physics, 4th Edition} - by Randall D. Knight \cite{knight2015physics}
\item \textbf{Mathematical Models in Biology, 1st Edition} - by Leah Edelstein-Keshet \cite{edelstein2005mathematical}
\item \textbf{Fundamentals of Physics, 10th Edition} - by David Halliday, Robert Resnick, and Jearl Walker \cite{halliday2013fundamentals}
\item \textbf{Mathematical Methods in the Physical Sciences, 3rd Edition} - by Mary L. Boas \cite{boas2006mathematical}
\item \textbf{Heat and Mass Transfer: Fundamentals and Applications, 5th Edition} - by Yunus A. Çengel and Afshin J. Ghajar \cite{cengel2014heat}
\item \textbf{Materials Science and Engineering: An Introduction, 8th Edition} - by William D. Callister Jr. \cite{callister2022fundamentals}
\item \textbf{Fluid Mechanics in SI Units, 8th Edition (2017)} - by Frank M. White \cite{white2012fluid}
\item \textbf{University Physics with Modern Physics, 14th Edition} - by Hugh D. Young and Roger A. Freedman \cite{young2010university}
\item \textbf{Analytical Dynamics, 1st Edition} - by Haim Baruh \cite{baruh2015applied}
\item \textbf{Engineering Electromagnetics, 8th Edition} - by William H. Hayt Jr. and John A. Buck \cite{hayt2001engineering}
\item \textbf{Modern Physics, 2nd Edition} - by Kenneth S. Krane \cite{krane2019modern}
\item \textbf{Introduction to Quantum Mechanics, 2nd Edition} - by David J. Griffiths \cite{griffiths2018introduction}
\item \textbf{Quantum Mechanics: Concepts and Applications, 2nd Edition} - by Nouredine Zettili \cite{zettili2009quantum}
\item \textbf{Classical Mechanics: An Undergraduate Text, 2006 Edition} - by John R. Taylor \cite{taylor2006classical}
\item \textbf{Introduction to Special Relativity, 1989 Edition} - by Robert Resnick \cite{resnick1989special}
\item \textbf{The Physics of Sound, 3rd Edition} - by Richard E. Berg and David G. Stork \cite{berg2012physics}
\item \textbf{Spacetime and Geometry: An Introduction to General Relativity, 2004 Edition} - by Sean M. Carroll \cite{carroll2004introduction}
\end{itemize}
\clearpage
\onecolumn
\subsection{Solution Refinement Example}
To provide full transparency regarding the dataset curation pipeline, this section provides illustrative `before' and `after' examples of solution refinement. As stipulated in Appendix \ref{app:dataset}, the `Before' text represents the human-originated solution scraped from educational forums. The `After' text is the final version included in \physicsevals following syntactic and stylistic refinement by Gemini 2.5 Pro. This example demonstrates that the model’s function was strictly limited to:
\begin{itemize}
    \item \textbf{Syntactic Correction:} Fixing grammar, typos, and standardizing language.
    \item \textbf{Formatting Consistency:} Converting disparate equation formats into uniform \LaTeX.
    \item \textbf{Structural Organization:} Clearly delineating the existing reasoning into explicit, sequential steps.
\end{itemize}
Critically, the core scientific logic, the mathematical formulations, and the step-by-step reasoning path remain unaltered from the human-authored original, confirming that the LLM's role was solely that of a sophisticated editor, not a content generator.
\begin{tcolorbox}[colback=gray!5, colframe=black, title={Poorly formatted human-generated solution}]
\begin{lstlisting}[
    basicstyle=\ttfamily\small,
    breaklines=true,          % enable automatic line wrapping
    breakatwhitespace=false,  % wrap anywhere, not only at spaces
    columns=fullflexible      % makes wrapping behave better for monospaced fonts
]
"ground_truth": " 
01 
 Given data 
 The value of the principal quantum number is, n = 17 .
02 
 Significance of number of points of zero and maximum probability does a matter wave have 
 In this question, the number of points of zero probability that a matter wave have can be obtained by adding the number 1 to the principal quantum number. The number of points of maximum probability would be the same as the principal quantum number.
03 
 (a) Determination of the number of points of zero probability 
 The expression to calculate the number of points of zero probability that a matter wave has are- 
 N 
 0 
 = n + 1 
 Here, N 
 0 
  is the number of points of zero probability does its matter have. 
 Substitute all the known values in the above equation. 
 N 
 0 
 = 17 + 1 
 = 18 
 Thus, the number of points of zero probability that a matter wave has are 18 .
04 
 (b) Determination of the number of points of maximum probability 
 The expression to calculate the number of points of maximum probability that a matter wave has is written as, 
 N 
 m a x 
 = n 
 Here, N 
 m a x 
  is the number of points of maximum probability does its matter have. 
 For the given values, the above equation becomes- 
 N 
 m a x 
 = 17 
 Thus, the number of points of maximum probability that a matter wave has is 17 ."
\end{lstlisting}
\end{tcolorbox}
\begin{tcolorbox}[
  colback=gray!5,
  colframe=black,
  title={Polished Dataset Entry},
  sharp corners,
  boxsep=4pt,
  left=2pt, right=2pt, top=4pt, bottom=4pt
]
\begin{lstlisting}[
    basicstyle=\ttfamily\normalsize,
    breaklines=true,          % enable automatic line wrapping
    breakatwhitespace=false,  % wrap anywhere, not only at spaces
    columns=fullflexible      % makes wrapping behave better for monospaced fonts
]
{
    "Problem_ID": "5dcead6a-2738-4be4-a750-751a51a3675d",
    "problem": " An electron is trapped in a one-dimensional infinite potential well in a state with quantum number $n = 17$. How many points of (a) zero probability and (b) maximum probability does its matter wave have? ",
    "simplified_problem_statement": "An electron confined within an infinite potential well is in the quantum state $n=17$. Determine the number of locations where the probability of finding the electron is zero and the number of locations where this probability is maximum.",
    "category": "Quantum Mechanics and Atomic Physics",
    "soft_labels": [
        "infinite potential well",
        "quantum number",
        "probability",
        "matter wave",
        "nodes",
        "antinodes"
    ],
    "elaborated_solution_steps": "Certainly! As a Professor of Physics, I'd be happy to elaborate on the provided solution step-by-step.\n\n**Step 01: Given data**\nThe value of the principal quantum number is, n = 17.\n\n**Energy of the nth state:**\n$E_n = \\frac{n^2 \\pi^2 \\hbar^2}{2mL^2}$\n\n**Wavefunction:**\n$\\psi_n(x) = \\sqrt{\\frac{2}{L}}\\sin\\left(\\frac{n\\pi x}{L}\\right)$\n\n**Number of nodes:** $N_0 = n + 1$\n\n**Number of antinodes:** $N_{max} = n$",
    "final_answers_in_brief": [
        "Number of points of zero probability: 18",
        "Number of points of maximum probability: 17"
    ],
    "source": "Fundamentals of Physics, 10th Edition - Halliday, Resnick, Walker"
}
\end{lstlisting}
\end{tcolorbox}

\subsection{Further Information}
\label{sec:further_info}
\begin{table}[ht]
\centering
\resizebox{\linewidth}{!}{%
\begin{tabular}{lrrrr}
\toprule
\textbf{Statistic} & \textbf{problem\_difficulty} & \textbf{steps} & \textbf{problem\_tokens} & \textbf{solution\_tokens} \\
\midrule
Count & 19609 & 19609 & 19609 & 19609 \\
Mean & 5.720282 & 3.883523 & 98.815595 & 3830.767403 \\
Std Dev & 1.445578 & 1.776616 & 78.899039 & 2458.892157 \\
Min & 1.000000 & 0.000000 & 1.000000 & 311.000000 \\
25\% & 5.000000 & 3.000000 & 51.000000 & 2422.000000 \\
50\% (Median) & 6.000000 & 4.000000 & 81.000000 & 3115.000000 \\
75\% & 7.000000 & 5.000000 & 124.000000 & 4300.000000 \\
Max & 10.000000 & 23.000000 & 4380.000000 & 29931.000000 \\
\bottomrule
\end{tabular}
}
\caption{Descriptive statistics for problem dataset}
\label{tab:problem_stats}
\end{table}
\begin{table}
\centering
\resizebox{0.8\linewidth}{!}{%
\begin{tabular}{lll}
\toprule
\textbf{Problem Category} & \textbf{Total Problems} & \textbf{Test Set} \\
\midrule
Acoustics and Sound Engineering & 589 & 58 \\
Quantum Mechanics and Atomic Physics & 1677 & 155 \\
Thermodynamics and Heat Transfer & 2451 & 234 \\
Solid State Physics and Materials Science & 789 & 78 \\
Fluid Mechanics and Continuum Dynamics & 936 & 95 \\
Electromagnetism and Circuit Theory & 2791 & 259 \\
Optics and Wave Phenomena & 1301 & 132 \\
Classical Mechanics and Dynamics & 4103 & 443 \\
Nuclear and Particle Physics & 766 & 76 \\
Statistical Mechanics and Kinetic Theory & 171 & 18 \\
Astrophysics, Cosmology, and Space Science & 961 & 105 \\
Relativity and Gravitation & 656 & 68 \\
Mathematical Methods in Physics & 1991 & 193 \\
Biophysics and Chemical Physics & 148 & 18 \\
Environmental and Atmospheric Physics & 43 & 5 \\
Measurement, Instrumentation, and Estimation & 171 & 17 \\
Engineering and Applied Physics & 41 & 5 \\
Computational and Numerical Methods & 13 & 1 \\
Food Physics and Culinary Science & 11 & 2 \\
\bottomrule
\end{tabular}
}
\caption{\label{Dataset Classification}
Categories of problems and their amounts in the dataset
}
\end{table}
\begin{figure*}
    \centering
    \includegraphics[width=1\linewidth]{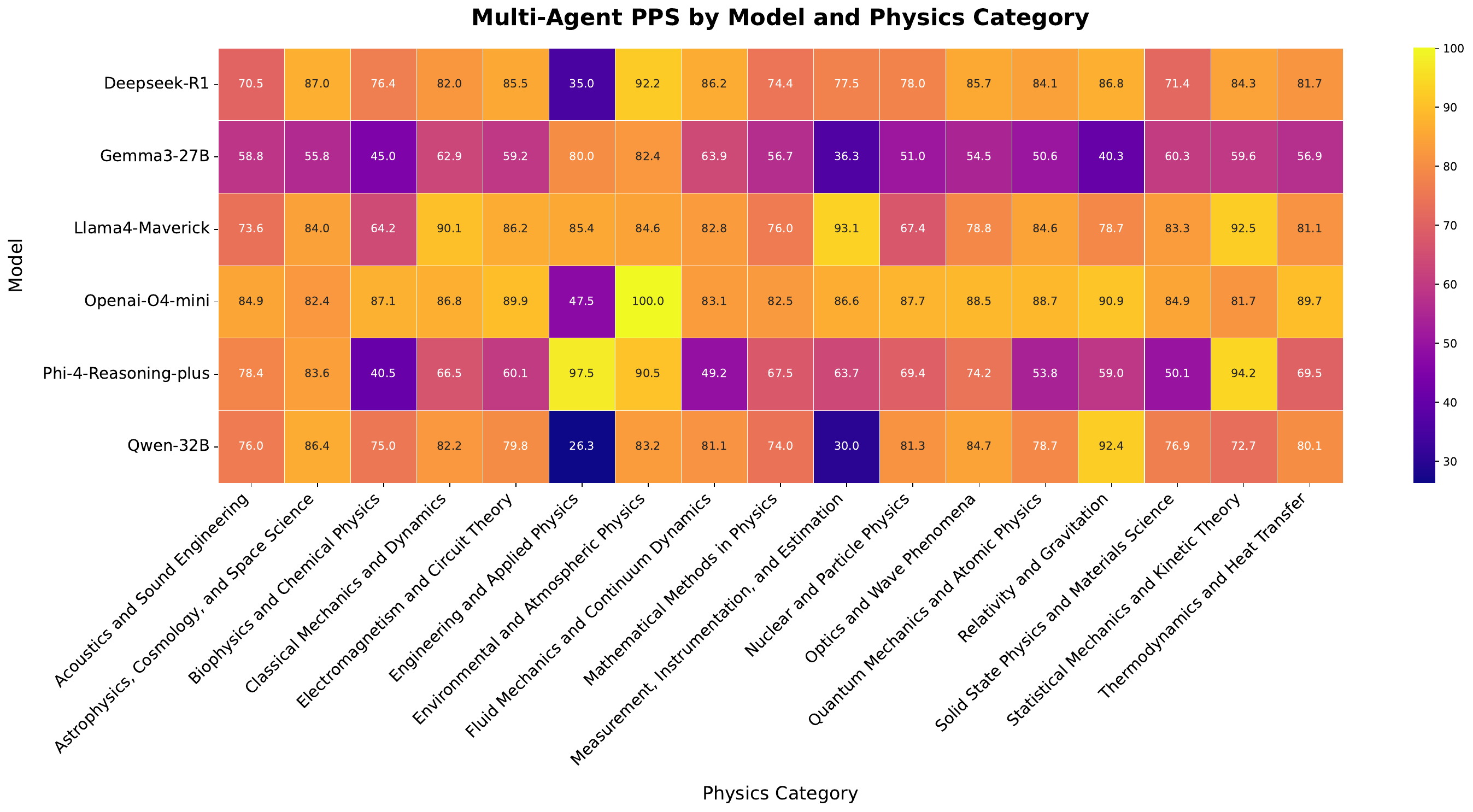}
    \caption{Multi-Agent PPS by Model and Physics Category}
    \label{fig:mar_pps_model_category}
\end{figure*}
\clearpage
\begin{tcolorbox}[colback=gray!5, colframe=black, title={Example from \physicseval}]
\noindent \textbf{Problem\_ID}

\texttt{46b6dfac-1f0c-4e23-9230-798ce854e963}

\noindent \textbf{problem}

 A long homogeneous resistance wire of radius \(r_{o}=\) \(5 \mathrm{~mm}\) is
being used to heat the air in a room by the passage of electric current. Heat
is generated in the wire uniformly at a rate of \(5 \times 10^{7} \mathrm{~W} /
\mathrm{m}^{3}\) \dots \dots Take the thermal conductivity of the wire to be \(k=6 \mathrm{~W} /
\mathrm{m} \cdot \mathrm{K}\).

\noindent \textbf{simplified\_problem\_statement}

A long wire with a radius of 5 mm generates heat uniformly at a rate of 5 x 10\^7 W/m\^3. The outer surface temperature of the wire is maintained at 180°C. Determine the temperature at a distance of 3.5 mm from the center of the wire, given that the thermal conductivity of the wire is 6 W/m·K.

\noindent \textbf{category}

Thermodynamics and Heat Transfer

\noindent \textbf{soft\_labels}

\begin{itemize}
        \item Heat Generation
        \item Steady-State Heat Transfer
        \item Radial Heat Conduction
        \item Boundary Conditions
        \item Thermal Conductivity
\end{itemize}

\noindent \textbf{elaborated\_solution\_steps}

Certainly! As a Professor of Physics, I'd be glad to elaborate on each step of the provided solution to ensure you understand the underlying physics and mathematical operations thoroughly. Let's break down each step:

**Step 01: The heat generation \dots \dots however needs correction for accurate temperature distribution.  But as instructed, I have elaborated on each step as provided in the solution, highlighting the physics and math, without altering the given equations.

\noindent \textbf{alternative\_solutions}

\begin{itemize}
        \item One could use a finite element method to solve the heat equation numerically, especially if the geometry or boundary conditions were more complex.
\end{itemize}

\noindent \textbf{problem\_difficulty}

6

\noindent \textbf{final\_answers\_in\_brief}

\begin{itemize}
        \item Temperature at r=3.5 mm is 178.86 °C
\end{itemize}

\noindent \textbf{steps}

6

\noindent \textbf{source}

Heat and Mass Transfer: Fundamentals and Applications, 5th Edition - by Yunus A. Çengel and Afshin J. Ghajar

\noindent \textbf{problem\_tokens}

171

\noindent \textbf{solution\_tokens}

11736
\end{tcolorbox}

\end{document}